\documentclass[10pt,twocolumn,letterpaper]{article}

\usepackage{cvpr}              % To produce the CAMERA-READY version
\usepackage{graphicx}
\usepackage{amsmath}
\usepackage{amssymb}
\usepackage{amsfonts}
\usepackage{booktabs}
\usepackage{epsfig}
\usepackage{multirow}
\usepackage{hyperref}
\hypersetup{colorlinks,allcolors=black}

\usepackage[hang, flushmargin]{footmisc}         %Problem line.
\usepackage{footnotebackref}
\usepackage[inline]{enumitem}
\usepackage{algorithm}
\usepackage{algpseudocode}
\usepackage{balance}
\usepackage[capitalize]{cleveref}
\crefname{section}{Sec.}{Secs.}
\Crefname{section}{Section}{Sections}
\Crefname{table}{Table}{Tables}
\crefname{table}{Tab.}{Tabs.}

\def\cvprPaperID{00027} % *** Enter the CVPR Paper ID here
\def\confName{CVPR}
\def\confYear{2022}

\begin{document}

\title{M2FNet: Multi-modal Fusion Network for Emotion Recognition in Conversation}

\author{Vishal Chudasama, Purbayan Kar, Ashish Gudmalwar, Nirmesh Shah, Pankaj Wasnik\thanks{Pankaj Wasnik is the corresponding author.}, Naoyuki Onoe \\ Media Analysis Group, Sony Research India, Bangalore, India\\
\tt\small\{vishal.chudasama1, purbayan.kar, ashish.gudmalwar, nirmesh.shah,\\ 
\tt\small pankaj.wasnik, naoyuki.onoe\}@sony.com}

\maketitle
%%%%%%%%% ABSTRACT
\begin{abstract}
Emotion Recognition in Conversations (ERC) is crucial in developing sympathetic human-machine interaction. In conversational videos, emotion can be present in multiple modalities, i.e., audio, video, and transcript. However, due to the inherent characteristics of these modalities, multi-modal ERC has always been considered a challenging undertaking. Existing ERC research focuses mainly on using text information in a discussion, ignoring the other two modalities. We anticipate that emotion recognition accuracy can be improved by employing a multi-modal approach. Thus, in this study, we propose a Multi-modal Fusion Network (M2FNet) that extracts emotion-relevant features from visual, audio, and text modality. It employs a multi-head attention-based fusion mechanism to combine emotion-rich latent representations of the input data. We introduce a new feature extractor to extract latent features from the audio and visual modality. The proposed feature extractor is trained with a novel adaptive margin-based triplet loss function to learn emotion-relevant features from the audio and visual data. In the domain of ERC, the existing methods perform well on one benchmark dataset but not on others. Our results show that the proposed M2FNet architecture outperforms all other methods in terms of weighted average F1 score on well-known MELD and IEMOCAP datasets and sets a new state-of-the-art performance in ERC.
\end{abstract}

%%%%%%%%% BODY TEXT
\section{Introduction}
Emotions are the unseen mental states that are linked to thoughts and feelings\cite{Poria2019Emotion}. In the absence of physiological indications, they could only be detected by human actions such as textual utterances, visual gestures, and acoustic signals. Emotion Recognition in Conversations (ERC) seeks to recognize the human emotions in conversations depending on their textual, visual, and acoustic cues. Recently, ERC has become an essential task in multimedia content analysis and moderation. It is a prominent trait to understand the nature of the interaction between users and the content. It has applications in various tasks, namely, AI interviews, personalized dialog systems, sentiment analysis, and understanding the user's perception of the content from the platforms like YouTube, Facebook, and Twitter\cite{Poria2019Emotion}.
\begin{figure}[t!]
    \centering
    \includegraphics[width = 0.5\textwidth, height = 0.2\textheight]{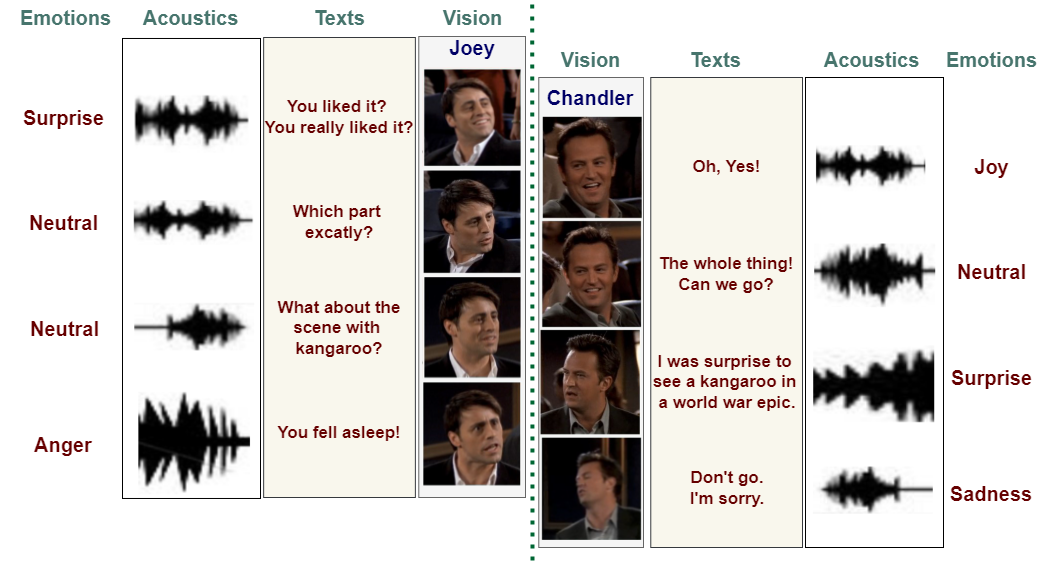}
    \caption{Multi-modal data as input}
    \label{fig:input1}
\end{figure}

In literature, we can see that many state-of-the-art methods adopt text-based processing to perform robust ERC \cite{kim2021emoberta, DBLP:journals/corr/abs-2010-02795}, such methods do not take into consideration the vast amount of information present in the acoustic and visual modalities. Since the ERC data mainly consists of all three modalities, i.e., text, visual, and acoustic, we hypothesize that the robust fusion of these modalities can improve the performance and robustness of the existing systems. A sample of emotional expressions in three different modalities is presented in Figure \ref{fig:input1} where the ERC system takes each modality as input and predicts the associated emotion.

In this paper, we propose a multi-modal fusion network (M2FNet) that takes advantage of the multi-modal nature of real-world media content by introducing a novel multi-head fusion attention layer. This layer combines features from different modalities to generate rich emotion-relevant representations by mapping the information from acoustic and visual features to the latent space of the textual features. In addition, we propose a new feature extractor model to extract the deeper features from the audio and visual contents. Here, we introduce a new adaptive margin-based triplet loss function, which helps the proposed extractor to learn representations more effectively. Additionally, we propose a dual network inspired from \cite{lee2019context} to combine the emotional content from the scene by taking into account the multiple people present in it. Furthermore, from the literature, we can see that state-of-the-art ERC methods perform well on one benchmark dataset, for example, IEMOCAP \cite{Busso2008IEMOCAPIE} while their performance degrades on more complex datasets like MELD \cite{poria2018meld}. This motivates us to propose a robust multi-modal ERC system. 

\begin{figure}[t!]
    \centering
    \includegraphics[width = 0.495\textwidth, height = 0.2\textheight]{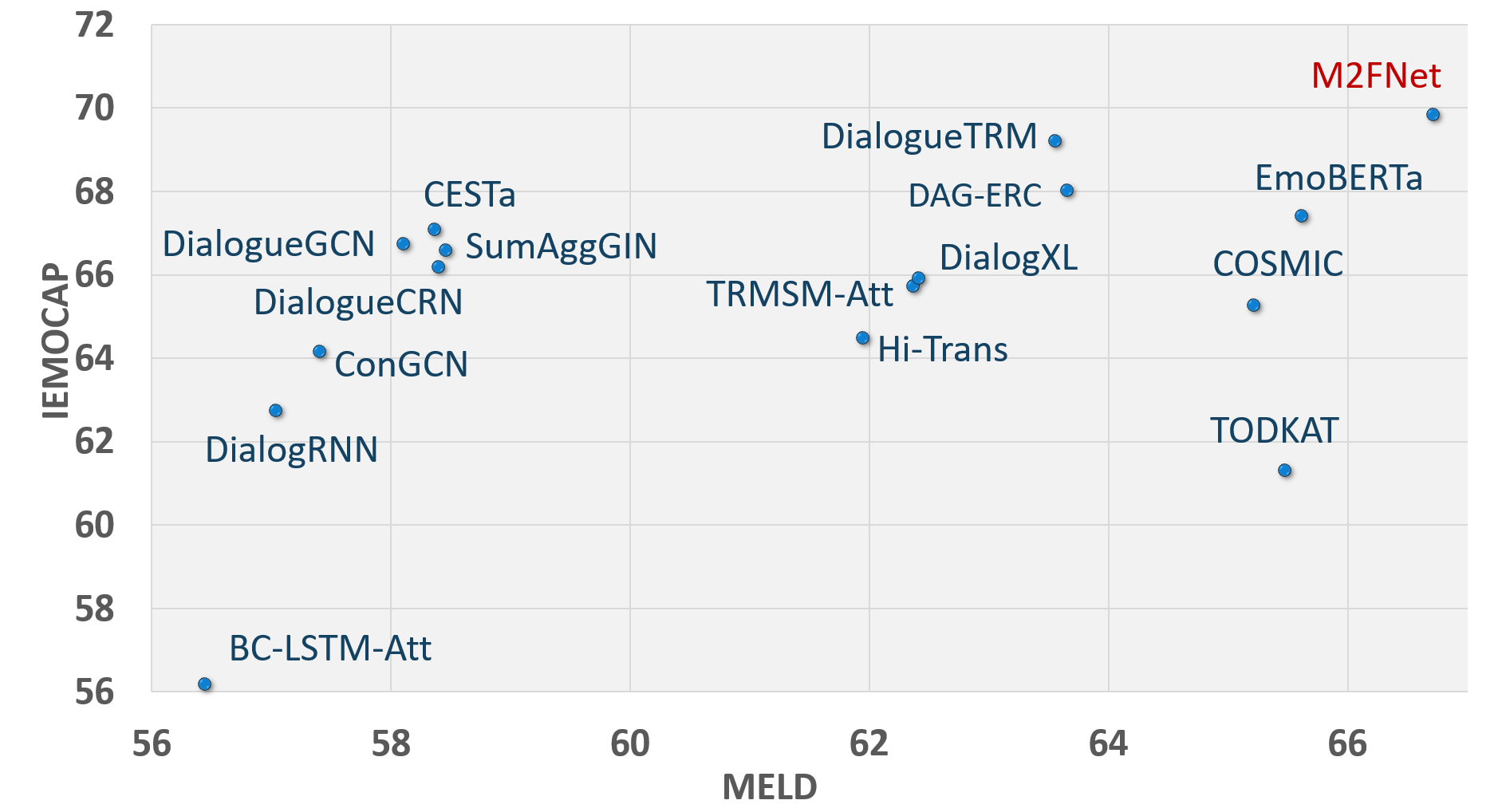}
    \caption{Quantitative analysis on MELD and IEMOCAP datasets in terms of weighted average F1 score.}
    \label{fig:input2}
\end{figure}

In order to verify the robustness of the proposed network, one experiment is carried out to compare its performance with existing text-based and multi-modal ERC methods. This comparison is visualized in Figure~\ref{fig:input2} where results are given in terms of weighted average F1 score on MELD \cite{Busso2008IEMOCAPIE} and IEMOCAP \cite{poria2018meld} datasets. Here, it can be observed that the proposed M2FNet model obtains a higher weighted average F1 score than other models. 
Followings are our major contributions:
\begin{itemize}
    \item A novel multi-modal fusion network called M2FNet is proposed for emotion recognition in conversation.
    \item A multi-head attention-based fusion layer is introduced, which aids the proposed system to combine latent representations of the different inputs.
    \item To extract deeper relevant features from audio and visual modality utterances, we introduce a new feature extractor model. 
    \item In the feature extractor model, we propose a new adaptive margin-based triplet loss function that helps the proposed model to learn emotion-relevant features.  
    \item To take advantage of the scene's emotional content, we also present a weighted face model that considers multiple people present in the scene.
    % \item Robustness of the proposed network is demonstrated on two benchmark datasets, where the proposed network outperforms many existing state-of-the-art ERC systems.
\end{itemize}

%-------------------------------------------------------------------------
\section{Related Works}
Emotion recognition in conversations (ERC) is different from traditional emotion recognition. Rather than treating emotions as static states, ERC involves emotional dynamics of a conversation, in which the context plays a vital role. Prior works on ERC mainly which use text, and audio features are proposed in \cite{lee2005toward, devillers2006real}. In the past few years, datasets with visual, acoustic and textual cues have been made publicly available \cite{Busso2008IEMOCAPIE, poria2018meld}. On these datasets, several deep learning methods are applied to recognize emotion. These techniques can be classified based on the type of data; either they merely utilize text or use multi-modal data (i.e. text, visual and audio). 
\subsection{Text-based methods}	
With the advent of the Transformer \cite{transformer}, the focus on text-based methods has recently increased. Due to the vast amount of information present in text data, current methods approach ERC as a purely text-based problem. Li \textit{et al.} \cite{li2020multi} use BERT \cite{devlin2018bert} to encode the individual sentences and then uses a  dialog level network for multitask learning on auxiliary tasks to generate better latent representations of the dialog as a whole. Furthermore, Li \textit{et al.} \cite{li-etal-2020-hitrans} build on this by incorporating transformer at the dialog end. Jiangnan \textit{et al.} \cite{li2020hierarchical} took this one step further by using the contextual representations from a BERT and transformer dialog network by designing three types of masks and utilizing them in three independent transformer blocks. The three designed masks learn the conventional context, Intra-Speaker, and Inter-Speaker dependency.
% In \cite{shen2020dialogxl}, authors modify the recurrence mechanism of XLNet from segment-level to utterance-level. Furthermore, they introduce dialog-aware self-attention to replace the vanilla self-attention in XLNet. 

In \cite{DBLP:journals/corr/abs-2010-02795}, Ghosal \textit{et al.} incorporates different elements of commonsense such as mental states, events, and causal relations to learn interactions between interlocutors participating in a conversation. Authors in \cite{ghosal2019dialoguegcn} and \cite{sheng-etal-2020-summarize} use Graph Neural networks to encode inter utterance and inter speaker relationships. Kim \textit{et al.} \cite{kim2021emoberta} model contextual information by simply prepending speaker names to utterances and inserting separation tokens between the utterances in a dialogue. To generate contextualized utterance representations, Wang \textit{et al.} \cite{wang-etal-2020-contextualized} uses LSTM-based encoders to capture self and inter-speaker dependency of interlocutors. A directed acyclic graph (DAG) based ERC was introduced by Shen \textit{et al.} in \cite{shen2021directed} which is an attempt to combine the strengths of conventional graph-based and recurrence-based neural networks. In \cite{zhu2021topic}, Zhu et al. propose a new model in which the transformer model fuses the topical and commonsense information to predict the emotion label. Recently, Song et al. \cite{emotionflow} proposed the EmotionFlow model, which encodes the user's utterances via concatenating the context with an auxiliary question, and then, a random field is applied to capture the sequential information at the emotion level. 
	
\subsection{Multi-modal Methods}
Prior literature on using previous utterances to provide context with respect to the utterance in hand has set the benchmark for dyadic conversations. In \cite{hazarika2018conversational, poria-etal-2017-context}, authors use previous utterances of both parties in a dyadic conversation and the contextual information from the same to predict the emotional state of any given utterance. Majumder \textit{et al.} \cite{majumder2018multimodal} build on this by separately modeling the uni-modal contextual information and then using a hierarchical tri-modal feature level fusion for obtaining a rich feature representation of the utterance. DialogueRNN \cite{DBLP:journals/corr/abs-1811-00405} tracks the contextual information of each speaker and the global state as separate entities. It uses the global and speaker's emotional context to produce accurate predictions. Zhang \textit{et al.} \cite{ijcai2019-752} introduced a model called ConGCN, which uses Graph Convolution networks on both audio and Text utterance features to model Speaker-Utterance and Utterance-Utterance relationships concurrently in a single network. Mao \textit{et al.} \cite{mao2020dialoguetrm} investigate the differentiated multi-modal emotional behaviors from the intra- and inter-modal perspectives. On a similar dataset, the CMU-Mosei methods like Loshchilov et al \cite{DBLP:journals/corr/abs-2006-15955} and Tsai et al \cite{tsai2019multimodal} use multi-head attention based fusion \cite{transformer} for multi-modal emotion recognition. \\

In most of the previous work, methods do not consider distinct facial features that play a significant role in determining the emotional context of the conversation. They use the frames as a whole entity but do not extract the essential part of the frame (i.e., face). Additionally, most of these methods do not have an active fusion strategy other than simple concatenation to take advantage of the wealth of the information present in the form of visual and acoustic data.

%------------------------------------------------------------------------
\section{Proposed Framework}
This section first introduces the problem statement and then provide details of the architecture design of the proposed framework briefly.
\subsection{Problem Statement}
A dialog consists of $k$ number of utterances ($U$) along with their respective labels ($Y$) arranged together with respect to time where each utterance is accompanied by it's respective video clip, speech segment and text transcript. Mathematically, a dialog for $k$ number of utterances can be formulated as follows:
\begin{align}
\label{eqn:eqlabel}
\begin{split}
\{U, Y\} = \{\{x^i = <x_t^i, x_a^i, x_v^i>, y^i\}| i\in[1,k]\},
\end{split}
\end{align}
Here, $x^i$ denotes the $i^{th}$ utterance made up of corresponding $x_t$ (text), $x_a$ (audio) and $x_v$ (visual) component, while $y_i$ indicates respective $i^{th}$ utterance's emotion label. The proposed network takes this data as input and assigns the right emotion to any given utterance. 

\subsection{Multi-modal Fusion Network: M2FNet}
\begin{figure}[t!]
    \centering
    \includegraphics[scale=0.24]{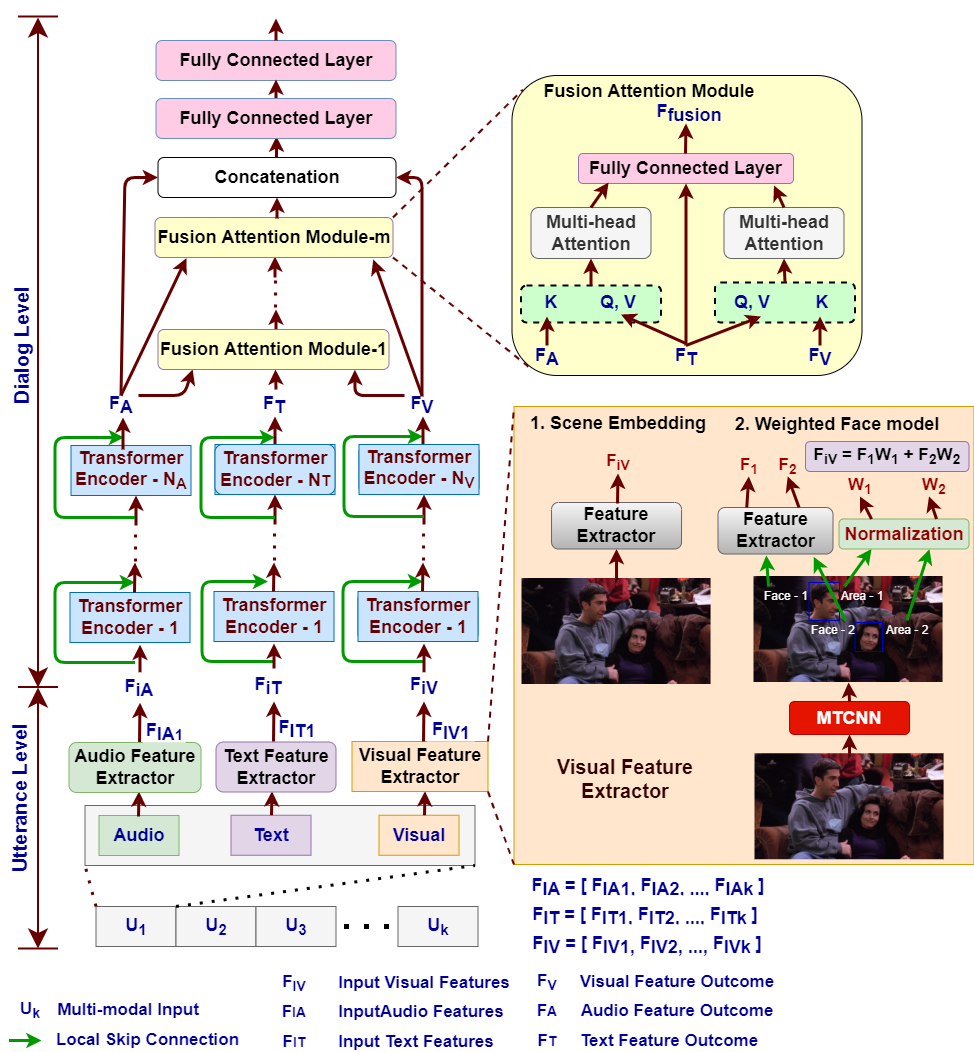}
    \caption{Network design of the proposed framework.}
    \label{fig:network}
\end{figure}

We propose a hierarchical framework called Multi-modal Fusion network (i.e., M2FNet) which is illustrated in Figure~\ref{fig:network}. The network is designed based on two levels of feature extraction: 
\begin{itemize}
    \item Utterance level feature extraction
    \item Dialog level feature extraction
\end{itemize}
Initially, the features are extracted by the utterance level module independently. Then, at the dialog level extraction network, the model learns to predict the right emotion for each utterance by using the contextual information from the dialog as a whole. In the subsequent subsections, we briefly discuss the design steps occurring in both utterance and dialog level feature extraction. 
\subsubsection{Utterance level feature extraction}
From Figure~\ref{fig:network}, one can see that there are $k$ number of utterances and each utterance is made up of $x_t$ (text), $x_a$ (audio) and $x_v$ (visual) content. In this step, the features from each modality are extracted for each utterance separately before passing to the dialog level feature extraction network. Each modality's input signals are passed through their corresponding feature extractors for generating their embeddings. %(features).

\textbf{Text:} In order to provide deeper inter utterance context, the text modality data (i.e., $x_t$) are passed through the Text Feature Extractor module. Here, we employ a modified RoBERTa model ($\phi_{M-RoBERTa}$) proposed by Kim \textit{et al.} \cite{kim2021emoberta} as feature extractor. Every utterance's $x_t$ is accompanied by its preceding and next utterance text separated by the separator token $<S>$. The modified RoBERTa model is fine-tuned on this transcript and the respective utterance's labels. The last layer activations $\{F_{IT}: F_{1,T},F_{2,T}......F_{k,T}$\} obtained from the modified RoBERTa model by passing the utterance's text data can be represented by:
\begin{equation}
    F_{IT}^i = \phi_{M-RoBERTa}(x_t^i)\ | \  i\in[1,k],\ \forall F_{IT} \in \mathbb{R}^{k\cdot D_T}.
\end{equation}
Here, $F_{IT}^i$ denotes $i^{th}$ utterance's embeddings and $D_T$ denotes the size of the embeddings of text utterance.

\textbf{Audio:} On the audio end, we introduce a new feature extractor model. %which is designed to leverage the importance of triplet loss function. 
The network design of the proposed feature extractor module is discussed briefly in Subsection \ref{sec:extractor}. Initially, the audio contents are transformed into 2D Mel Spectrogram in RGB format and then passed through the feature extractor model. 
% Figure~\ref{fig:audio} shows the generation process of Mel Spectrogram from the given audio signal. 
Here, the audio signal is first processed via different augmentation techniques like time warping and Additive White Gaussian Noise (AWGN) noise. Then the augmented signals are transformed into the corresponding Mel Spectrograms \cite{mel}. For computing the Mel Spectrogram, the Short Time Fourier transform (STFT) is used with the frame length of 400 samples (25 ms) and hop length of 160 samples (10ms). We also use 128 Mel filter banks to generate the Mel Spectrogram \cite{MelBank}. 

The proposed extractor takes the Mel Spectrograms (i.e., $x_a$) as input and generate the corresponding feature embeddings \{$F_{IA}: F_{1,A},F_{2,A}......F_{k,A}$\}. The functionality of the proposed audio feature extractor module can be mathematically expressed as,
\begin{equation}
    F_{IA}^i = \phi_{AFE}(x_{a}^i)\ | \  i\in[1,k],\ \forall F_{IA}  \in \mathbb{R}^{k\cdot D_A}
\end{equation}
where, $F_{IA}^i$ is the $i^{th}$ utterance's embeddings and $D_A$ indicates the size of embeddings of audio utterance and the $\phi_{AFE}$ denotes the function of introduced audio feature extractor module.

\textbf{Visual:} In order to extract rich emotion-relevant features from the visual signal, we propose a dual network inspired from \cite{lee2019context} that exploit not only human facial expression but also context information in a joint and boosting manner. For both tasks, we use our proposed extractor model (as discussed in Subsection \ref{sec:extractor}) and train on the CASIA webface database \cite{CASIA} to extract the deeper features from the visual image. %This trained extractor model is then performed on 15 successive frames of the utterance clip to extract the deeper features. 
%The proposed dual network architecture is visualized in Figure~\ref{fig:network}. 
Following are the details of steps involved in the dual network:
\begin{itemize}
    % \item We also use our proposed feature extractor model to extract the deeper features from the visual images. For that, we train our extractor model on CASIA webface database []. The details of the  Then the trained extractor model is used to encode the context information of the scene as a whole Then our trained feature extractor model is performed on 15 successive frames of the utterance clip. Following which the features are max pooled over the frame axis to obtain the scene embeddings of the utterance. 
    \item To encode the context information of the scene as a whole our trained feature extractor model is performed on 15 successive frames of the utterance clip. Following which the features are max pooled over the frame axis to obtain the scene embeddings of the utterance. 
    
    \item To extract facial emotion-related features from the same 15 successive frames of the utterance clip, we propose a weighted Face Model (i.e., as visualized in Figure~\ref{fig:network}) which works as follows: 
    \begin{itemize}
        \item Given a frame, it is passed through a Multi-task Cascaded Convolutional Network (MTCNN) \cite{7553523} to detect the faces present in the frame. This returns the bounding box of each face along with its confidence. Then each of the respective faces is passed through our trained feature extractor model to obtain emotion-relevant features of each respective face. Now, the areas of each bounding box accompanying the faces are normalized to bring their values between 0 and 1. Following this, a weighted sum is performed using the features of each face and their respective normalized areas to obtain the facial emotion feature of a frame.
        \item The same process is followed for each of the 15 frames, and similarly, upon extraction, the features are max-pooled over the frame axis to obtain the facial features of the utterance.
    \end{itemize}
\end{itemize}

Upon extraction of features from each network, the scene embeddings are concatenated with the facial features to obtain a more comprehensive representation of the visual data in an utterance. Finally, this visual feature extractor's output $\{F_{IV}: F_{1,V},F_{2,V}......F_{k,V}$\} can be formulated as:
\begin{equation}
    F_{IV}^i = \phi_{WF}(x_{v}^i)\ | \  i\in[1,k],\ \forall F_{IV} \in \mathbb{R}^{k\cdot D_V}
\end{equation}
where, $F_{IA}^i$ is the $i^{th}$ utterance's embeddings and $\phi_{WF}$ denotes the function operation of weighted face model while $D_V$ is the size of the feature embedding of the visual utterance. 
Finally, these embeddings (i.e., text, acoustic and visual) are then sent to the dialog level feature extractor as an input to learn the correct prediction of emotions for each utterance.

\subsubsection{Dialog level feature extraction}
The design diagram at dialog level feature extraction of the proposed M2FNet model is illustrated in Figure~\ref{fig:network}. Each modality embeddings (i.e., $F_{IT}, F_{IA}$ and $F_{IV}$) are passed through their corresponding network with a variable stack of transformer encoders \cite{transformer} to learn the inter utterance context. The number of transformer encoders for text, audio, and visual modality is denoted by $N_T, N_A$, and $N_V$, respectively. We also employ a local skip connection between each encoder to prevent the model from ignoring the lower-level features. The corresponding feature maps obtained from the encoders can be mathematically expressed as:
\begin{align}
\label{eqn:eqlabel1}
\begin{split}
 F_T^i = Tr_{N_T}(...(Tr_{N_2}(Tr_{N_1}(F_{IT}^i)))),
\\
 F_A^i = Tr_{N_A}(...(Tr_{N_2}(Tr_{N_1}(F_{IA}^i)))),
 \\
 F_V^i = Tr_{N_V}(...(Tr_{N_2}(Tr_{N_1}(F_{IV}^i)))), 
 \\ 
 where,\  i\in[1,k].
\end{split}
\end{align}
Here, $Tr$ is the operation function of the transformer encoder.

The corresponding feature maps associated with the text, visual and audio (i.e., $F_T, F_V, F_A$) are passed to a novel Multi-Head Attention Fusion module that helps the network in incorporating visual and acoustic information. The network architecture of the attention fusion module is also depicted in Figure~\ref{fig:network}. Here, the text features $F_T$ are used as input to fusion module as \textbf{Query} (Q) and \textbf{Value} (V) for the multi-head attention operation, and then the visual $F_V$ and acoustic $F_A$ features for the dialog are used as \textbf{Key} (K) in order to modulate the attention given to each utterance at any time-step. Hence, each individual modality is now mapped to the text vector space, and the respective features are concatenated and passed to a fully connected layer which outputs a vector $\in \mathbb{R}^{k\cdot D_T}$. The output of the fusion layer is passed through the next fusion layer along with the previous $F_A$ and $F_V$ feature maps. Here, the $m$ number of multi-head attention fusion layers stacked together in order to generate the final feature outcome ($F_{fusion}$) as demonstrated below:
\begin{align}
\begin{split}
F_{fusion_1}^i = \Phi_1(F_A^i, F_T^i, F_V^i),\qquad\qquad\qquad\qquad\ \  \\
F_{fusion_m}^i = \Phi_m(...(\Phi_2(F_A^i, F_{fusion_1}^i, F_V^i))),\ \\
where,\  i\in[1,k].
\end{split}
\end{align}
In the above equation, $\Phi$ indicates the learning function of the proposed Multi-Head Attention Fusion layer. 
The main difference in our Fusion strategy compared to the previous work using Multi-Head Attention is that our strategy involves changing the key across modalities while keeping the Query and Value the same to better modulate inter utterance attention and incorporate inter-modal information.

The feature outcome of the last multi-head attention fusion layer (i.e., $F_{fusion_m}$) is concatenate with visual and acoustic feature maps (i.e., $F_V$ and $F_A$) as expressed below:
\begin{equation}
    F_{final}^i = Concate(F_{fusion_m}^i, F_A^i, F_V^i)
\end{equation}

Finally, we append two fully connected layers (FC) which generates the desired predicted output (i.e., $Y_p^i = <y_1^i, y_2^i, ..., y_p^i>$, where,  $i\in[1,k]$) of our proposed system.
\begin{equation}
    y_p^i = FC(FC(F_{final}^i)).
\end{equation}

\subsection{Feature Extractor Module}\label{sec:extractor}
\begin{figure}[t!]
    \centering
    \includegraphics[width = 0.485\textwidth, height = 0.15\textheight]{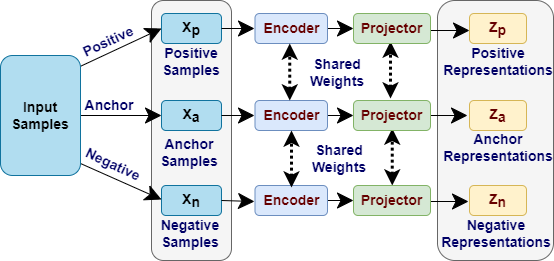}
    \caption{Network design of the proposed Extractor network.}
    \label{fig:extractor}
\end{figure}
In order to fetch deep features from audio and visual contents, we propose a new feature extractor model and the same is illustrated in Figures~\ref{fig:extractor}. 
The proposed extractor is designed based on triplet network to leverage the importance of triplet loss function \cite{facenet}.
Initially, the anchor, positive and negative samples have been generated as suggested in \cite{facenet} for audio and visual modalities. Then these samples are passed through encoder network followed by projector module.
% The proposed extractor model consists encoder network followed by projector module. 
Here, we use standard ResNet18 \cite{resnet18} as a backbone of the encoder network while the projector consists a linear fully connected layer which project the embedding of encoder network to desired representations (i.e., $Z = [z_1, ... z_N] \in \mathcal{R}^{N \times d}$) composed of $N$ representations with dimension $d$. 

The proposed extractor model is trained using weighted combination of three loss function i.e., adaptive margin triplet loss (i.e., $L_{AMT}$), covariance loss (i.e., $L_{Cov}$) and variance loss (i.e., $L_{Var}$) functions. It can be expressed as follow:
\begin{equation}\label{eq:extractor}
    L_{FE} = \lambda_1 \cdot L_{AMT} + \lambda_2 \cdot L_{Cov} + \lambda_3 \cdot L_{Var},
\end{equation}
where, $\lambda_1$, $\lambda_2$ and $\lambda_3$ are weighting factors that controls the distribution of different loss functions.

In \cite{facenet}, authors design the triplet loss function used to learn good representations of faces based on anchor, positive and negative samples. 
%We utilize this approach to learn deeper features from audio modality contents. 
Here, authors have used a fixed margin value in their triplet loss function that helps to separate out the representations of positive and negative samples.
%will prevent the loss from being zero when the positive is closer to the anchor but is very close to the negative. 
However, in some cases where the positive or negative samples have the same distance with the anchor or the positive sample is only a bit closer to the anchor than the negative sample, the loss would be zero, and there would be no correction even though it should still be pulling the positive sample closer and pushing the negative sample away from the anchor. To overcome this issue, we propose a triplet loss function based on a adaptive margin value. This can be mathematically written as 
\begin{equation}
L_{AMT} = D_{s}^{a,p} - \frac{D_{s}^{a,n}+D_{s}^{p,n}}{2} + m_{AM}.  
\end{equation}
Here, $D_{s}^{a,p}$, $D_{s}^{a,n}$ and $D_{s}^{p,n}$ denotes the euclidean distance based similarity metric between representations of anchor and positive, anchor and negative, positive and negative samples, respectively. $m_{AM}$ is the adaptive margin which is calculated based on similarity and dissimilarity measures as
\begin{align}
\begin{split}
    m_{AM} &= m_{AM}^{sim} + m_{AM}^{dissim}\\
    m_{AM}^{sim} &= 1 + \frac{2}{e^{4\cdot D_{s}^{a,p}}}\\
    m_{AM}^{dissim} &= 1 + \frac{2}{e^{-4\cdot D_{s}^{a,n}+4}}
\end{split}
\end{align}

In addition, we also utilize the variance loss function proposed by Bardes et al. \cite{vicreg} which helps the proposed model to tackle the mode collapse issue. 
% This loss function defines the variance regularization term as a hinge loss on the standard deviation of the representations along batch dimension.
Mathematically, the variance loss function can be presented as,
\begin{align}
\begin{split}
    L_{Var} &= \sum_{k = 1}^3 L_{Var} (Z_k); \quad Z_k = Z_a, Z_p, Z_n\\
    L_{Var} (Z_k) &= \frac{1}{d}\sum_{j=1}^{d}1 - \sqrt{Var(Z_{:,j})+\epsilon}
\end{split}
\end{align}
Here, $Var(Z)$ (i.e., $\frac{1}{N-1} \sum_{i=1}^{N} (Z^i - \Hat{Z})^2$) denotes the variance obtained from the corresponding representations, while $\Hat{Z}$ is mean of the corresponding representation.

To decorrelate the different dimensions of the representations, we adopt the covariance loss function \cite{vicreg} and the the same can be expressed mathematically for representation as
\begin{align}
\begin{split}
L_{Cov} &= \sum_{k = 1}^3 L_{Cov} (Z_k); \quad Z_k = Z_a, Z_p, Z_n\\
L_{Cov} (Z_k) &= \frac{1}{d} \sum_{i\ne j}Cov(Z_k)^T_{i,j},
\end{split}
\end{align}
where, $Cov (Z) = \frac{1}{N-1} \sum_{i=1}^{N} (Z^i - \Hat{Z})(Z^i - \Hat{Z})^T$ indicates the covariance matrix of corresponding representation. 
%-------------------------------------------------------------------------
\section{Experimental Analysis and Discussion}

\subsection{Dataset}
For fair comparison with state-of-the-art methods, we evaluate our proposed network (M2FNet) on Interactive Emotional Dyadic Motion Capture (IEMOCAP) \cite{Busso2008IEMOCAPIE}, and Multimodal EmotionLines Dataset (MELD) \cite{poria2018meld} benchmark datasets. The statistics of these datasets are reported in Table \ref{Tab:statics} and details are provided in Section 1 of the supplementary material. Both IEMOCAP and MELD are multimodal datasets with textual, visual, and acoustic data. 
\begin{table}[t!]
\centering
\caption{Statistics of the testing benchmark datasets: MELD and IEMOCAP}\label{Tab:statics}
 \resizebox{0.4\textwidth}{!}{
\begin{tabular}{lcccc}
\toprule
\textbf{Statistics}      & \multicolumn{2}{c}{\textbf{MELD}}          & \multicolumn{2}{c}{\textbf{IEMOCAP}}\\ \toprule
%\multirow{2}{*}{\textbf{\begin{tabular}[c]{@{}l@{}}Dataset \\ Splitting\end{tabular}}} &  &  &  &  \\
Splitting & \textbf{\# Dialog} & \textbf{\# Utterance} & \textbf{\# Dialog} & \textbf{\# Utterance} \\
Train                              & 1098               & 9989                  & 100                & 4778                  \\
Dev                                & 114                & 1109                  & 20                 & 980                   \\
Test                               & 280                & 2610                  & 31                 & 1622                  \\ \midrule
\begin{tabular}[c]{@{}l@{}}No. of \\ Classes\end{tabular}  & \multicolumn{2}{c}{7}                      & \multicolumn{2}{c}{6}                      \\ \bottomrule
\end{tabular}}
\end{table}

\textbf{MELD:} The MELD \cite{poria2018meld} is a multimodal and multiparty dataset containing more than 1,400 conversations and 13,000 utterances from the Friends TV series. The utterances are annotated with one of the seven emotion labels (anger, disgust, sadness, joy, surprise, fear, and neutral). We use the pre-defined train/val split provided in the MELD dataset. The details of this dataset are given in Table \ref{Tab:statics}.

\textbf{IEMOCAP:} The IEMOCAP database \cite{Busso2008IEMOCAPIE} is an acted, multimodal and multi-speaker database consisting of videos of dyadic sessions having approximately 12 hours of audiovisual data with text transcriptions. Each video contains a single conversation, which is segmented into multiple utterances. Each utterance is annotated with one of six emotion labels, i.e., happy, sad, neutral, angry, excited, and frustrated.  The database statistics are given in Table \ref{Tab:statics}. % and Fig.~\ref{fig:statics}. 
We randomly select 10\% of training conversations as evaluation split for computing the hyperparameters.

\subsection{Training Setups and Hyper-parameter Tuning}
All experiments are carried out using a single NVIDIA GeForce RTX 3090 card. We adopt AdamW \cite{loshchilov2017decoupled} as the optimizer with an initial learning rate 5e-4 with $L_2$ weight decay ranges between 5e-4 and 5e-5. 
%We use an early stopping strategy on the weighted average F1-score of the validation set, with a patience of 5-20 epochs. 
Dropout is used with a rate between 0.4 and 0.5. The number of encoder layers in each modality's encoder (i.e., $N_T, N_A, N_V$) is tuned using a greedy scheme and set to 1 and 5 for MELD and IEMOCAP validation datasets, respectively. The number of multi-head attention fusion layers is set to 5 (i.e., $= m$) for both dataset. %Each modality has a variable number of encoders and attention heads, both of which are tuned separately, and these hyperparameters are combined together. 
The proposed M2FNet framework is trained using the categorical cross-entropy on each utterances softmax output for each of \textbf{M} dialogs and their \textbf{k} utterances. Section 2 of the supplementary material provides training and validation performance details.
\begin{equation}
    Loss=-\frac{1}{M \cdot k} \cdot\sum_{i=1}^M\sum_{j=1}^k\sum_{t=1}^C y^{i,j,t}\cdot\log(y_{p}^{i,j,t}).
\end{equation}

% The training performance of the proposed model is discussed in Section 2 in the supplementary material.

To extract deeper features from audio and visual contents, we introduce a new feature extractor model. Here, ResNet18 \cite{resnet18} is used as encoder module while the projector module consists a fully connected layer which projects the embeddings of encoder network to desired representations (i.e., $Z$). Here we set the number of representations as $Z = 300$. For audio task, the extractor model is trained on Mel Spectrograms obtained from the corresponding audio signals while in case of visual feature extraction, it is trained on well-known CASIA webface database \cite{CASIA}. Here, the extractor model is trained using the loss function mentioned in Equation no. \ref{eq:extractor}
in which the weighting factors $\lambda_1, \lambda_2$ and $\lambda_3$ are set to 20, 5 and 1, respectively. The proposed extractor model is trained upto 60 and 100 epochs for audio and visual task, respectively using Adam optimizer with learning rate of 1e-4 and decay rate of 1e-6. 

We mainly employ weighted average F1 score as evaluation metric due to its suitability to test with imbalance dataset. Additionally, we present our results in terms of classification accuracy to evaluate the model performance.

\begin{table}[t!]
\caption{Ablation Studies based comparison to validate the impact of each modality.}
\label{abl:modality}
\resizebox{0.5\textwidth}{!}{
\begin{tabular}{lccccc}
\hline
\multirow{2}{*}{Models} & \multirow{2}{*}{Remarks} & \multicolumn{2}{c}{IEMOCAP} & \multicolumn{2}{c}{MELD} \\ \cline{3-6} 
 &  & Accuracy & \begin{tabular}[c]{@{}c@{}}Weighted \\ Average F1\end{tabular} & Accuracy & \begin{tabular}[c]{@{}c@{}}Weighted \\ Average F1\end{tabular} \\ \hline
Only Audio & --- & 26.56 & 21.79 & 49.04 & 39.63 \\
Only Visual & --- & 20.39 & 13.10 & 45.63 & 32.44 \\
Visual + Audio & Concat & 35.12 & 31.35 & 48.35 & 35.74 \\
Only Text & --- & 66.30 & 66.20 & 67.24 & 66.23 \\
Text + Audio & Concat & 66.52 & 66.48 & 67.80 & 66.32 \\
Text + Visual & Concat & 66.64 & 66.67 & 67.81 & 66.35 \\
Text + Visual + Audio & Concat & 67.16 & 67.12 & 67.28 & \textbf{66.81} \\
Text + Visual + Audio & Fusion & \textbf{69.69} & \textbf{69.86} & \textbf{67.85} & 66.71 \\ \hline
\end{tabular}}
\end{table}

% \usepackage{multirow}
\begin{table}[t!]
\caption{Ablation Studies based comparison to validate the impact of dual network.}
\label{abl:dual}
\resizebox{0.5\textwidth}{!}{
\begin{tabular}{lcccc}
\hline
\multirow{2}{*}{Models} & \multicolumn{2}{c}{IEMOCAP} & \multicolumn{2}{c}{MELD} \\ \cline{2-5} 
 & Accuracy & \begin{tabular}[c]{@{}c@{}}Weighted\\ Average F1\end{tabular} & Accuracy & \begin{tabular}[c]{@{}c@{}}Weighted\\ Average F1\end{tabular} \\ \hline
Scene Embeddings & 67.65 & 67.70 & 65.23 & 64.23 \\
Weighted Face Embeddings & 67.32 & 67.28 & 66.37 & 65.67 \\
Proposed & \textbf{69.69} & \textbf{69.86} & \textbf{67.85} & \textbf{66.71} \\ \hline
\end{tabular}}
\end{table}

\begin{table}[t!]
\caption{Ablation Studies based comparison to validate the impact of no. of transformer encoders.}
\label{abl:encoders}
\resizebox{0.5\textwidth}{!}{
\begin{tabular}{ccccc}
\hline
\begin{tabular}[c]{@{}c@{}}No. of \\ Transformer Encoders \\ (i.e., $N_A = N_T = N_V$)\end{tabular} & \multicolumn{2}{c}{IEMOCAP} & \multicolumn{2}{c}{MELD} \\ \cline{2-5} 
 & Accuracy & \begin{tabular}[c]{@{}c@{}}Weighted \\ Average F1\end{tabular} & Accuracy & \begin{tabular}[c]{@{}c@{}}Weighted\\ Average F1\end{tabular} \\ \hline
1 & 69.32 & 69.36 & \textbf{67.85} & \textbf{66.71} \\
2 & 69.07 & 69.21 & 67.16 & 65.82 \\
3 & 69.32 & 69.44 & 66.86 & 66.06 \\
4 & 69.13 & 69.22 & 67.20 & 66.06 \\
5 & \textbf{69.69} & \textbf{69.86} & 67.47 & 66.38 \\
6 & 69.13 & 69.26 & 67.24 & 65.48 \\ \hline
\end{tabular}}
\end{table}

\begin{table}[t!]
\caption{Ablation Studies based comparison to validate the impact of multi-head attention fusion layers.}
\label{abl:fusion}
\resizebox{0.5\textwidth}{!}{
\begin{tabular}{ccccc}
\hline
\begin{tabular}[c]{@{}c@{}}No. of  Attention \\ Fusion Layers\end{tabular} & \multicolumn{2}{c}{IEMOCAP} & \multicolumn{2}{c}{MELD} \\ \cline{2-5} 
 & Accuracy & \begin{tabular}[c]{@{}c@{}}Weighted\\ Average F1\end{tabular} & Accuracy & \begin{tabular}[c]{@{}c@{}}Weighted\\ Average F1\end{tabular} \\ \hline
1 & 67.22 & 67.34 & 66.63 & 66.11 \\
2 & 68.08 & 68.19 & 67.47 & 66.50 \\
3 & 69.13 & 69.25 & 67.47 & 66.65 \\
4 & 69.01 & 68.95 & 67.59 & 66.51 \\
5 & \textbf{69.69} & \textbf{69.86} & \textbf{67.85} & \textbf{66.71} \\
6 & 68.95 & 69.08 & 66.82 & 65.32
\end{tabular}}
\end{table}

\subsection{Ablation studies}
To better understand the contribution of different modules in the proposed M2FNet model, we have conducted several ablation studies on both IEMOCAP and MELD datasets. 
% Here, we train the proposed network on different scenarios such as with and/or without Text, Video, and Audio, with/without Fusion attention layer, and with an only scene and weighted face embeddings. 
The corresponding results are compared in terms of accuracy and weighted average F1 scores for MELD and IEMOCAP testing datasets.

To validate the impact of each modality, we train the proposed network with and/or without Text, Video, and Audio as input and without using the proposed Fusion module. From Table \ref{abl:modality}, one can observe that concatenation of multi-modal input with all three modalities obtain higher accuracy and weighted average F1 score than other scenarios such as using only one or two modalities.
Furthermore, our fusion mechanism helps to enhance the accuracy by 2.53\% and 0.57\% for IEMOCAP and MELD datasets, respectively. However, in the case of weighted average F1 score, it obtains slightly inferior performance for the MELD dataset while improving it by 2.74\% for the IEMOCAP dataset. 

On the visual end, we utilize scene and weighted face embeddings. For understanding the importance of both embeddings, two more experiments have been carried out where the proposed network with individual embedding has been trained, and the corresponding results are given in Table \ref{abl:dual}. From the table, it can be observed here that the weighted face model or the scene encoding network on their own do not improve the results; however, when the network can access both, it significantly improves the results. This shows that both the context from the scene and the people in the scene are equally important for emotion recognition. 

We also observe the effect of transformer encoders in the proposed framework. In these experiments, we set same number of transformer encoders (i.e., $N_A = N_V = N_T$) for each modality and observe its effect for different numbers. The corresponding results are presented in Table \ref{abl:encoders} where it is observed that $N_A = N_V = N_T = 1$ gives best performance for MELD dataset while the $N_A = N_T = N_V = 5$ setting helps the proposed framework to obtain higher performance for IEMOCAP testing dataset.

In the proposed model, we have set $m = 5$ number of Multi-Head Attention Fusion modules. To validate this, we train the proposed model with different numbers of the Multi-Head Attention Fusion modules and observe the corresponding accuracy and weighted average F1 score. This analysis is demonstrated in Table \ref{abl:fusion} where one can observe that the proposed model with five Multi-Head Attention Fusion modules (i.e., m = 5) obtains higher quantitative measures on both datasets.
\subsection{Model Performance}
\begin{figure}[t!]
    \centering
\includegraphics[width = 0.485\textwidth, height = 0.17\textheight]{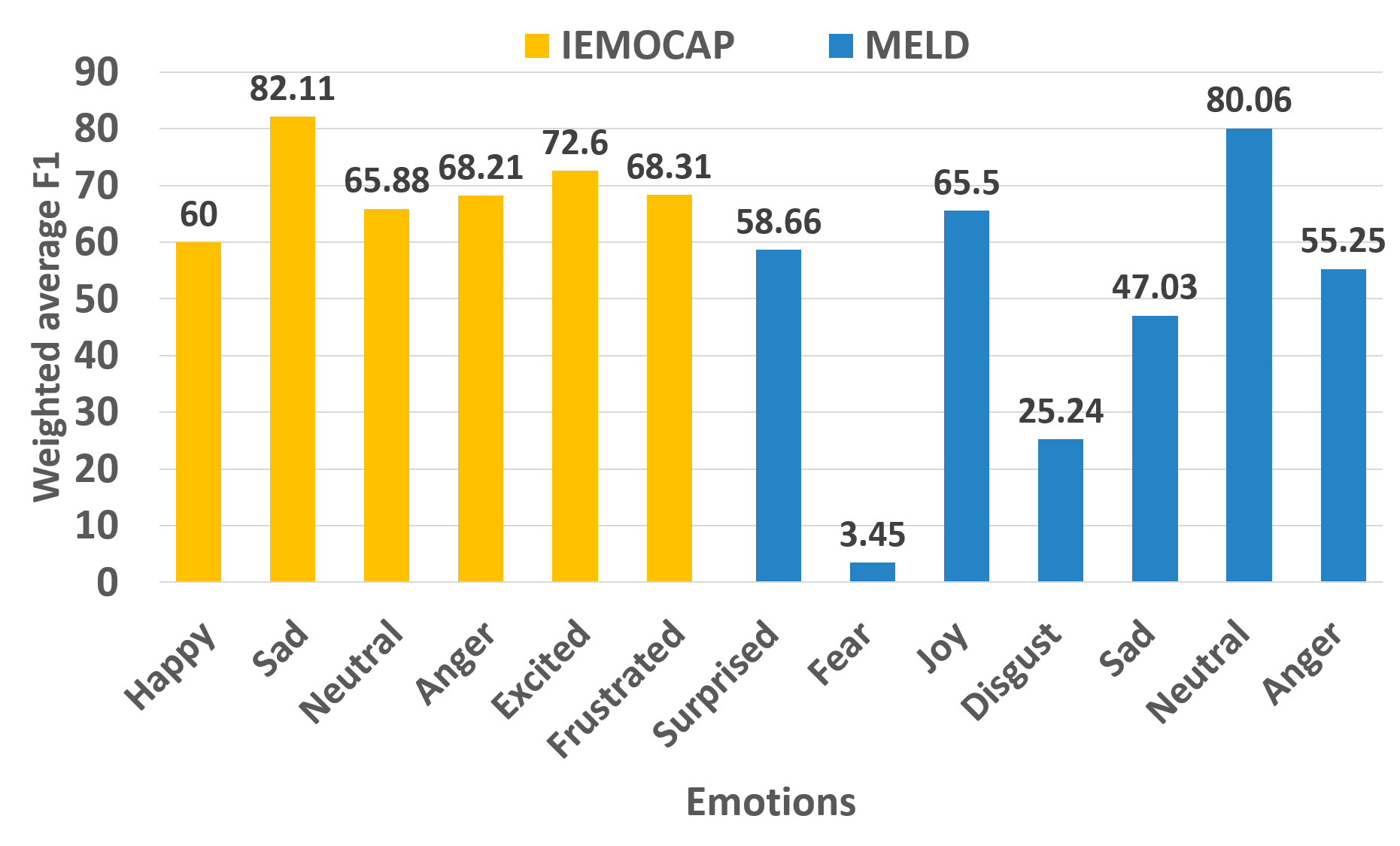} \caption{Predictions made by the network on the MELD and IEMOCAP test sets}
    \label{fig:prediction}
\end{figure}

Figure~\ref{fig:prediction} presents the performance of our model in terms of the weighted average F1 score of different emotions. Here, we can see that our model has obtained the highest F1 score of 82.11\% for the Sad emotion of the IEMOCAP dataset. Similarly, for MELD dataset, it obtained highest score for Neutral emotion. 

\begin{table}[t!]
	\centering
	\caption{Quantitative comparison with text-based state-of-the-art methods in terms of weighted average F1 score.} \label{tab:comparison-text}
	\resizebox{0.36\textwidth}{!}{
	\begin{tabular}{lccc}
		\toprule
%		\cmidrule(r){1-3}
		Name & Year  & MELD & IEMOCAP\\
		\midrule
        BERT+MTL \cite{li2020multi}           & 2020  & 61.90  &   --- \\
		Hi-Trans \cite{li-etal-2020-hitrans} & 2020 & 61.94 & 64.50  \\
		TRMSM-Att \cite{li2020hierarchical}  & 2020 & 62.36 &65.74 \\
		DialogXL \cite{shen2020dialogxl}   &2020   & 62.41 &65.94 \\
		COSMIC \cite{DBLP:journals/corr/abs-2010-02795} &2020 & 65.21 &65.28 \\
		SumAggGIN \cite{sheng-etal-2020-summarize}      &2020  & 58.45 & 66.61 \\
		DialogueGCN \cite{ghosal2019dialoguegcn}   & 2019    & 58.10 & 66.76 \\
		CESTa \cite{wang-etal-2020-contextualized}    & 2020  & 58.36 & 67.10            \\
		DialogueCRN \cite{hu2021dialoguecrn}  & 2021     &  58.39 & 66.20              \\
		EmoBERTa \cite{kim2021emoberta}    & 2021  & 65.61 & 67.42 \\
		DAG-ERC \cite{shen2021directed}   & 2021   & 63.65 & 68.03 \\
		TODKAT \cite{zhu2021topic}*   & 2021   & 65.47 & 61.33 \\
		EmotionFlow \cite{emotionflow}   & 2022   & 66.50 & --- \\
		\midrule
		%\textbf{\textit{M2FNet$_{\, Text}$}}     & 2021  & 66.41 & \textbf{68.05}    \\
		\textbf{\textit{M2FNet}} &2022 & \textbf{66.71} & \textbf{69.86} \\
		\bottomrule
		\multicolumn{4}{l}{\begin{tabular}[l]{@{}l@{}}* The performance of TODKAT model is updated by\\ its author in their official repository.\end{tabular}}
	\end{tabular}}
\end{table}
\subsection{Comparative Analysis}
%This paper aims to propose a robust ERC model that can perform well on both benchmark  datasets. 
To validate the robustness of the proposed network, we compare our proposed network with state-of-the-art text-based ERC systems in terms of weighted average F1 score and the same is presented in Table \ref{tab:comparison-text}. Here, one can notice that the proposed network has a state-of-the-art performance by obtaining superior quantitative results than previous methods on both datasets (i.e., 0.21\% higher than previous best EmotionFlow \cite{emotionflow} model on MELD dataset while 1.83\% higher than that of the previous best DAG-ERC \cite{shen2021directed} model on IEMOCAP testing dataset). 

\begin{table}[t!]
	\centering
	\caption{Quantitative comparison with multimodal-based state-of-the-art methods on MELD and IEMOCAP datasets. Here, top two performances are highlighted with bold font texts.}\label{tab:comparison-multimodal}
	%\vspace{10pt}
	\resizebox{0.48\textwidth}{!}{
\begin{tabular}{lcccc}
\toprule
\multicolumn{1}{c}{\textbf{Name of Model}} & \multicolumn{2}{c}{\textbf{MELD}}     & \multicolumn{2}{c}{\textbf{IEMOCAP}}        \\ \cline{2-5} 
\multicolumn{1}{c}{}                                                                                & \multicolumn{1}{c}{Accuracy} & \multicolumn{1}{c}{\begin{tabular}[c]{@{}c@{}}Weighted \\ Average F1\end{tabular}} & \multicolumn{1}{c}{Accuracy} & \multicolumn{1}{c}{\begin{tabular}[c]{@{}c@{}}Weighted \\ Average F1\end{tabular}} \\ \hline
BC-LSTM-Att \cite{poria-etal-2017-context} &57.50     &     56.44    &56.32   & 56.19    \\
DialogRNN \cite{DBLP:journals/corr/abs-1811-00405}  &59.54     &     57.03    &63.40   & 62.75    \\
ConGCN \cite{ijcai2019-752}     & ---    &     57.40    & 64.18  & 64.18   \\
Xie at al. \cite{crossmodel}  &   65.00  &     64.00    &  --- & ---   \\
DialogueTRM \cite{mao2020dialoguetrm} & 65.66    &     63.55    &  68.92 & 69.23  \\
\textbf{M2FNet} &    \textbf{67.85} & \textbf{66.71}    & \textbf{69.69}  & \textbf{69.86}  \\ \bottomrule
\end{tabular}}
\end{table}

When compared with the existing multi-modal methods, our proposed M2FNet network shows a substantial improvement compared to state-of-the-art multi-modal ERC methods. The obtained results are presented in Table \ref{tab:comparison-multimodal} where one can notice that the proposed M2FNet model obtains 2.19\% and 2.71\% higher accuracy and weighted average F1 score on MELD dataset than that of previous best performance. Similarly, it set 0.77\% and 0.63\% higher accuracy and weighted average F1 sore than that of previous best DialogueTRM model \cite{mao2020dialoguetrm} on IEMOCAP testing dataset.

\section{Limitations}
Our model sometimes gets confused and miss-classifies similar or close emotions such as Frustration and Anger, Happy and Excited. We can also observe that for highly imbalance data, our model misclassifies many emotions as the emotion with higher number of data samples. For example, many emotions are overwhelmingly predicted as \textit{Neutral} for MELD dataset.

\section{Conclusion}
In this paper, we propose a robust multi-modal fusion network called M2FNet for the task of Emotion Recognition in Conversation. In M2FNet, we propose a multi-head fusion attention module that helps the network to extract rich features from multiple modalities. A new feature extractor model is introduced in the proposed design to learn the audio and visual features effectively. Here, a new adaptive margin triplet loss function is introduced, which helps the extractor module to learn representations effectively. A new weighted face model is proposed in our framework to learn the rich facial features. Detailed analysis shows that encoding both scene and face-related information is essential for emotion recognition. Similarly, we observed that multi-modal fusion is necessary to leverage information from multiple modalities present in an utterance. Finally, our experiments validate the robustness of the proposed network quantitatively on both benchmark datasets. 

%%%%%%%%% REFERENCES
%\balance
\newpage
\balance
{\small
\bibliographystyle{ieee_fullname}
\bibliography{egbib}
}
\end{document}